\documentclass[10pt,twocolumn,letterpaper]{article}

\usepackage[pagenumbers]{cvpr} 

\usepackage{graphicx}
\usepackage{amsmath}
\usepackage{amssymb}
\usepackage{booktabs}
\usepackage[normalem]{ulem}
\usepackage[table,xcdraw]{xcolor}
\usepackage[linesnumbered,ruled,vlined]{algorithm2e}
\usepackage{makecell} 
\usepackage{multirow}

\usepackage[pagebackref,breaklinks,colorlinks]{hyperref}

\usepackage[capitalize]{cleveref}
\crefname{section}{Sec.}{Secs.}
\Crefname{section}{Section}{Sections}
\Crefname{table}{Table}{Tables}
\crefname{table}{Tab.}{Tabs.}

\def\cvprPaperID{*****} 
\def\confName{CVPR}
\def\confYear{2023}

\begin{document}

\title{GPr-Net: Geometric Prototypical Network\\ for Point Cloud Few-Shot Learning}

\author{Tejas Anvekar\\
KLE Technological University\\ Center of Excellence in Visual Intelligence \\
{\tt\small anvekartejas@gmail.com}
\and
Dena Bazazian\\
University of Plymouth\\ Faculty of Science and Engineering\\
{\tt\small dena.bazazian@plymouth.ac.uk}
}

\maketitle

\begin{abstract}
In the realm of 3D-computer vision applications, point cloud few-shot learning plays a critical role. However, it poses an arduous challenge due to the sparsity, irregularity, and unordered nature of the data. Current methods rely on complex local geometric extraction techniques such as convolution, graph, and attention mechanisms, along with extensive data-driven pre-training tasks. These approaches contradict the fundamental goal of few-shot learning, which is to facilitate efficient learning. To address this issue, we propose GPr-Net (Geometric Prototypical Network), a lightweight and computationally efficient geometric prototypical network that captures the intrinsic topology of point clouds and achieves superior performance. Our proposed method, IGI++ (Intrinsic Geometry Interpreter++) employs vector-based hand-crafted intrinsic geometry interpreters and Laplace vectors to extract and evaluate point cloud morphology, resulting in improved representations for FSL (Few-Shot Learning). Additionally, Laplace vectors enable the extraction of valuable features from point clouds with fewer points. To tackle the distribution drift challenge in few-shot metric learning, we leverage hyperbolic space and demonstrate that our approach handles intra and inter-class variance better than existing point cloud few-shot learning methods. Experimental results on the ModelNet40 dataset show that GPr-Net outperforms state-of-the-art methods in few-shot learning on point clouds, achieving utmost computational efficiency that is $170\times$ better than all existing works.  The code is publicly available at \href{https://github.com/TejasAnvekar/GPr-Net}{https://github.com/TejasAnvekar/GPr-Net}.
\end{abstract}

\begin{figure} [t]
\begin{center}
    \centering
    \captionsetup{type=figure}
    \includegraphics[width=\linewidth]{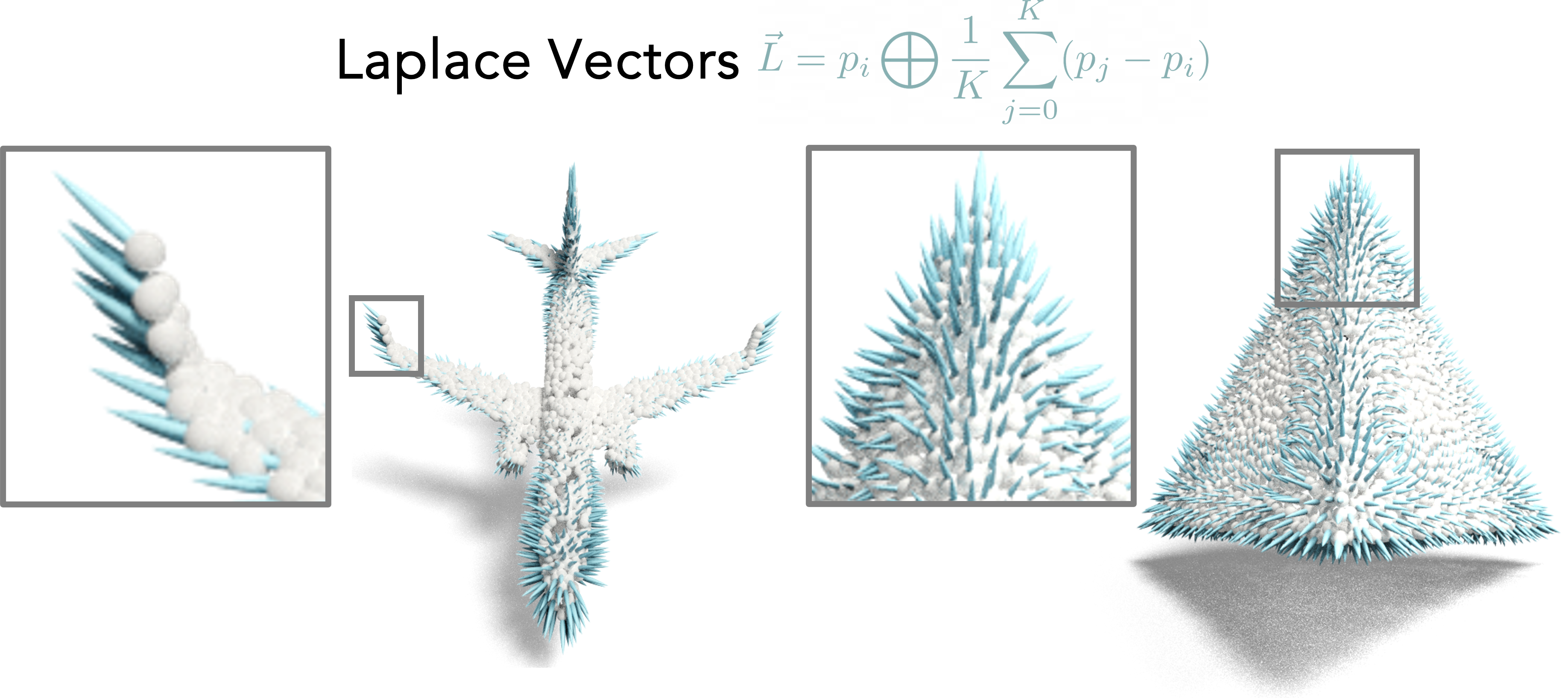}
    \captionof{figure}{We demonstrate Laplace vectors, a simple yet effective geometric signature that captures the statistics of group deviations, facilitating  the abstraction of edges and corners in point clouds. The efficacy of Laplace vectors is highlighted in the visualization of an airplane and a tetrahedron. In the airplane, Laplace vectors capture the uni-directed high group deviation at the ends of the wings, indicating a sudden change or an edge in local topology. Similarly, the tetrahedron exhibits high and uniform group deviation at its ends, indicating a corner.}
    \label{fig:teaser}
\end{center}%
\end{figure}

\section{Introduction}
\label{sec:intro}

The domain of computer vision has witnessed a remarkable surge in the significance of 3D data processing, with point cloud data emerging as a prominent representation obtained via real-time acquisition using LiDAR scanners. Point cloud object classification plays a critical role in several applications such as indoor SLAM~\cite{indoor}, robotics~\cite{robo}, and autonomous vehicles~\cite{selfdriving}, facilitating efficient navigation and decision-making. Deep learning-based techniques~\cite{pointnet}~\cite{HyCoRe} have revolutionized 3D point cloud classification by enabling the extraction of representative features from shape projections~\cite{mvcnn} or raw points, thereby enhancing performance compared to traditional handcrafted feature-based methods~\cite{HC1}~\cite{HC2}.

Despite the recent progress in geometric deep learning, the need for large amounts of labeled training data remains a significant challenge, both in terms of cost and practicality~\cite{pointnetpp}. While self-supervised approaches~\cite{3D-GAN}~\cite{PointCapsNet}, data augmentation~\cite{pointaug}~\cite{patchaug}, and regularisation~\cite{HyCoRe} techniques have helped alleviate the aforementioned issue, they may not perform well on new tasks or unseen classes without sufficient labeled training data. This has led to a growing demand for methods that enable geometric deep networks to quickly adapt to novel settings with limited labeled data, much like humans who can learn new concepts with only a few examples by drawing on prior knowledge and inductive bias\cite{inductive-bias}.

To address this challenge, few-shot learning (FSL) techniques~\cite{Prototypical_Network}~\cite{relational-Networks}~\cite{MAML} have shown remarkable progress in 2D visual understanding tasks such as image classification~\cite{HIE}, object detection~\cite{FSL-IOBD}, and semantic segmentation~\cite{FSL-SS}. However, FSL on 3D data is still in its nascent stages and presents unique challenges. Previous approaches to 3D-FSL~\cite{SS-FSL}~\cite{Enrich-features}~\cite{cia} have focused on determining the best FSL algorithm, network design, and deep learning methodology, often relying on complex pre-training tasks or intricate deep learning modules. These approaches may not effectively capture the human-inspired characteristics that researchers aim to incorporate, leading to limited generalization.

Towards addressing the aforementioned challenges, we introduce a novel 3D-FSL approach, the Geometric Prototypical Network (GPr-Net), which leverages geometric priors to achieve fast and efficient point cloud few-shot learning. Unlike conventional approaches that rely on complex pre-training~\cite{SS-FSL} procedures or sophisticated deep learning modules~\cite{cia}, GPr-Net is engineered to transfer geometric prior knowledge directly to novel tasks with minimal training. To capture these valuable geometric priors for 3D-FSL, we propose Intrinsic Geometry Interpreters++ (IGI++), which efficiently captures the local intrinsic topology of the point cloud using the IGI features inspired by the VG-VAE~\cite{Vg-vae}. Additionally, we propose Laplace vectors to extract abstract information about the edges and corners present in point clouds. The coherently combined intrinsic and Laplace vectors of IGI++ provide a comprehensive representation of the crucial geometric properties for few-shot learning on point clouds as shown in Figure~\ref{fig:teaser}. Furthermore, we address the distribution shift in prototypical networks by mapping our geometric priors to the Hyperbolic metric. Extensive experiments on the ModelNet40 dataset demonstrate the superiority of GPr-Net in few-shot learning on point clouds compared to state-of-the-art methods. GPr-Net achieves up to $170\times$ fewer parameters that facilitate faster performance and a 5\% increase in accuracy compared to related works.

Our contributions can be summarized as:
\begin{itemize}
    \item We propose GPr-Net: a lightweight Geometric Prototypical Network designed for fast and efficient point cloud few-shot learning.
    \item We propose an Intrinsic Geometry Interpreters++ (IGI++) to cohere intrinsic and high-frequency geometric signatures of a point cloud which comprises the following modules: 1) an Intrinsic Geometric Interpreter (IGI) to efficiently capture the local topology of the point cloud; 2) our proposed novel Laplace vectors to capture the abstraction of edges and corners in point clouds.
    \item We propose employing the Hyperbolic / poincar\'e metric to mitigate the challenge of distribution shift in prototypical networks.
    \item We demonstrate the impact of our derived geometric signatures on ModelNet40 and outperforms existing state-of-the-art few-shot learning techniques by 5\% in accuracy with $170\times$ fewer parameters.
\end{itemize}

\begin{figure*}
    \centering
    \includegraphics[width=0.95\linewidth]{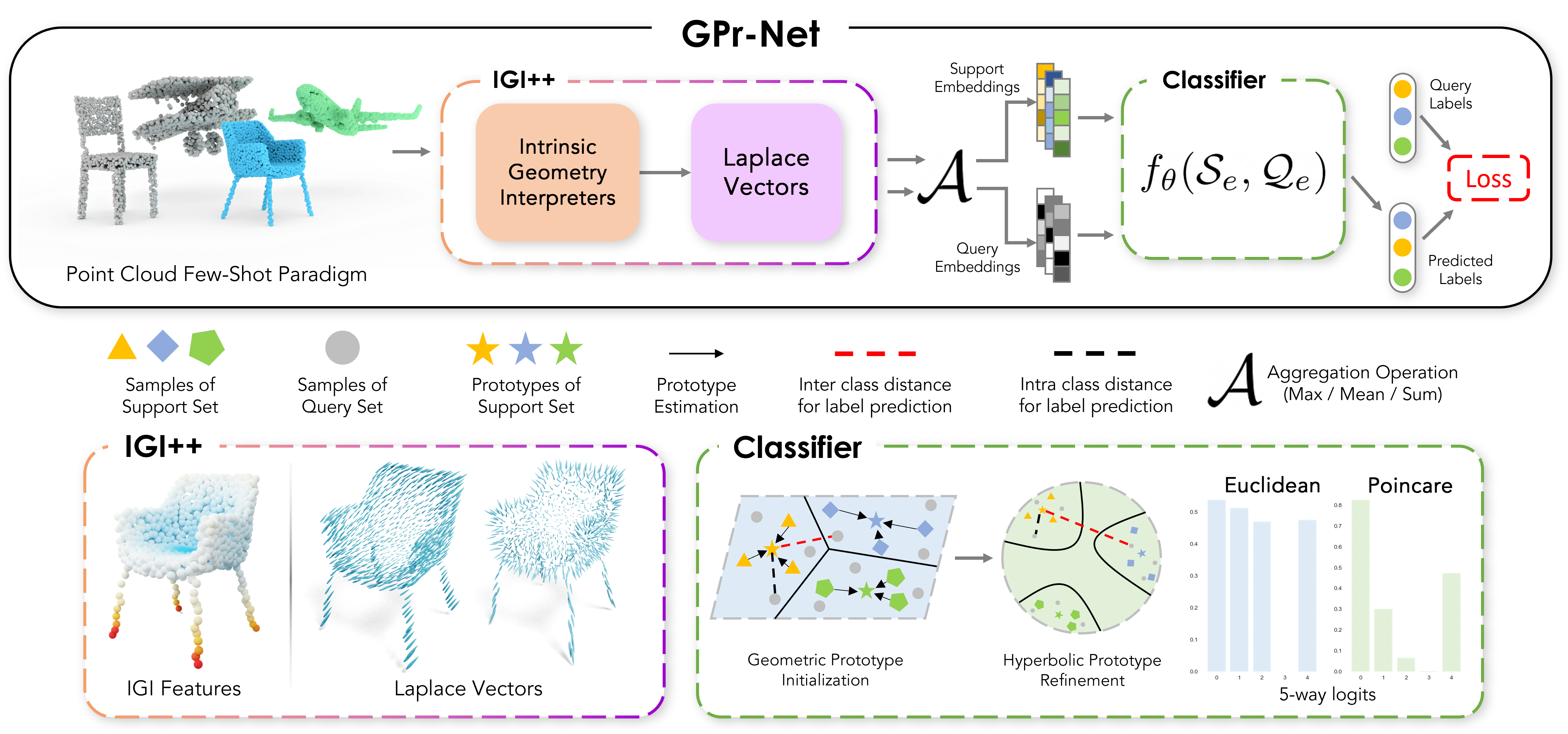}
    \caption{We present an overview of the proposed GPr-Net framework, which processes point clouds in a few-shot episodic paradigm using the proposed IGI\cite{Vg-vae} and Laplace vectors to generate geometric feature sets. These features are then mapped to a higher dimensional permutation invariant feature using the symmetric operation $\mathcal{A}$ and a single Multilayer Perceptron (MLP) $f_{\theta}$. The Prototypical network $f_{\theta}$, utilizes the support and query geometric embeddings $\vec{L}(\Psi(x_{s})) = \mathcal{S}_{e}$ and $\vec{L}(\Psi(x_{q})) = \mathcal{Q}_{e}$ to predict few-shot labels. To overcome the distribution drift challenge in Prototypical Networks, we employ the Hyperbolic Distance of Euclidean. For more details, please refer to Section \ref{sec:metrci_learning}.}
    \label{fig:gpr}
\end{figure*}

\section{Related Works}
\label{sec:RW}

\noindent \textbf{Point Cloud Analysis} has been revolutionized by deep learning models for 3D point cloud classification by allowing us to learn more intricate and representative features. Unlike traditionally handcrafted methods~\cite{HC1}~\cite{HC2}, these models can learn these features without any human intervention. There are two types of deep learning methods: projection-based and point-based networks. Projection-based~\cite{mvcnn} networks first transform irregular points into a structured representation such as voxel~\cite{VoxelNet} or lattices and then use standard convolution neural networks to extract view-wise or structural features. However, they may encounter explicit information loss or higher memory consumption. Point-based methods have become more popular, exemplified by the likes of PointNet~\cite{pointnet} and other approaches like PointNet++~\cite{pointnetpp} and PointCNN \cite{PointCNN}, and DGCNN~\cite{dgcnn}, Point-MLP~\cite{PointMLP}, HyCoRe~\cite{HyCoRe}, EDC-Net~\cite{EDC}, DCG-Net~\cite{DCG}, which utilize convolution or graph-based networks to achieve state-of-the-art performance. While these deep learning methods require a significant amount of annotated data, their generalization capabilities on novel classes during training may be limited. This limitation could potentially be a subject of future research.\\
        
\noindent \textbf{Few-shot learning} (FSL) has emerged as a crucial field in machine learning that aims to overcome the limitations of traditional supervised learning methods, which require large labeled datasets to generalize to new tasks. To achieve this, several approaches have been proposed, including Prototypical Networks \cite{Prototypical_Network}, which introduced the concept of prototypes for few-shot classification, and Relation Networks \cite{relational-Networks}, which proposed a novel architecture that captures relations between different instances to improve accuracy. Model-Agnostic Meta-Learning (MAML) \cite{MAML} takes a meta-learning approach to few-shot classification, learning an initialization of the model that can be quickly adapted to new tasks with only a few labeled examples. Recent research has tackled the challenge of 3D point cloud learning with limited training data. Sharma et al.~\cite{SS-FSL}. explore feature representation through self-supervision, while LSSB \cite{LSSB} aim to learn a discriminative embedding space for 3D model multi-view images. However, authors in  Enrich-Features \cite{Enrich-features}  proposed a novel few-shot point cloud classification paradigm that effectively combines current fully supervised methods. This approach utilizes feature fusion and channel-wise attention to improve feature learning accuracy. These works represent important strides in addressing the challenge of 3D point cloud learning with limited data. 

\noindent \textbf{Hyperbolic Metric Learning} embeds hierarchical structures with low distortion~\cite{hsurvey}. It has been used for non-Euclidean manifolds in various representation learning frameworks. Early on, it was used for natural language processing. Hyperbolic neural network~\cite{HNN} layers have been shown to be better than Euclidean ones, and hyperbolic variants have been explored for images and graphs~\cite{HGCNN}. Euclidean embeddings are insufficient for complex visual data. Hyperbolic Image Embeddings~\cite{HIE} address this by capturing hierarchical relationships with negative curvature, improving few-shot classification accuracy on benchmarks like miniImageNet~\cite{mini} and CUB~\cite{cub}. HyCoRe~\cite{HyCoRe} introduced a new method for using hyperbolic space embeddings to capture the part-whole hierarchy of 3D objects in point clouds. This approach significantly improves the performance of point cloud classification models. To the best of our knowledge, no research has explored using hyperbolic representations for few-shot learning of point clouds, despite their inherent hierarchical structure and ability to mitigate distribution drift of Prototypical networks. There exists a need for learning hyperbolic embeddings to capture the compositional nature of 3D objects with can facilitate the capture tree-like geometric hierarchy of the data, making them a superior prior for 3D-FSL.

Our method for point cloud few-shot learning introduces a novel approach that utilizes geometric signatures and Hyperbolic space to improve performance. It is distinguished from existing methods by its lightweight, fast, and pragmatic nature, requiring only a few episodes to train the challenging few-shot classification task.

\section{GPr-Net}
\label{sec:Metho}
    We present GPr-Net, a lightweight Geometric Prototypical Network designed for fast and efficient point cloud few-shot learning. By leveraging intrinsic geometric features, GPr-Net captures abstract information necessary for superior few-shot learning. Our proposed Intrinsic Geometry Interpreters++ (IGI++) extracts fundamental features like local topology, edges, and corners, while a single fully connected layer maps aggregation of geometric features to a higher dimensional point for episodic few-shot classification. Furthermore, we enhance our model's performance by incorporating Hyperbolic space that yields sharp logits for few-shot learning as depicted in Figure~\ref{fig:gpr}. Unlike previous methods, GPr-Net relies on statistical geometric features and is trained using a few-shot paradigm to simulate real-life scenarios.

\subsection{Notations and Strategies}
    Let $P$ denote point cloud such that $P = \{p_{1}, ... p_{n}\}$ where $p_{i} \in \mathbb{R}^{d}$ and $n$ represents total number of points. Inspired by the episodic paradigm of the Few-Shot Classification (FSL) task on Images \cite{HIE} we incorporate a similar algorithm with minimal changes for point cloud FSL. The train set $D_{train}$ and test set $D_{test}$ are designed such that the categories of  $D_{train} \cap D_{test} = \emptyset$. \\
    
    FSL is optimized for episodes containing pair of  $K$-ways and $N$-shots of the support set  and query  set. $N_\mathcal{S}$ samples are drawn from $N_C$ at randomly selected $K$ categories to form the support set $\mathcal{S}~=~\{( P^{1}_{s}, y^{1}_{s} ),... (P^{N_\mathcal{S} \times K}_{s}, y^{N_\mathcal{S} \times K}_{s} )\}$. The remaining $N_\mathcal{Q}$ form the query set $\mathcal{Q}~=~\{( P^{1}_{q}, y^{1}_{q} ),... (P^{N_\mathcal{Q} \times K}_{q}, y^{N_\mathcal{Q} \times K}_{q} )\}$. The goal is to predict $y_{q}^{i}$ via a model $f_{\theta}(\mathcal{S},\mathcal{Q})$ by only utilizing labels support set $y_{s}^{i}$.
    \

\subsection{Network Design}
    We advocate using Prototypical Networks~\cite{Prototypical_Network} as a method of choice for point cloud few-shot learning due to its simplicity and remarkable generalization of metric learning. Prototypical Networks large rely on network design as it plays a key role in initializing representation metrics that significantly enhances the performance of few-shot learning. Our network design comprises two fundamental components: geometric feature extraction using proposed IGI++ and metric learning with a single fully connected MLP, both of which synergistically facilitate point cloud few-shot learning. 


    \subsubsection{Intrinsic Geometry Interpreter++}
\begin{figure}[ht]
            \center
            \includegraphics[width=1.0\linewidth]{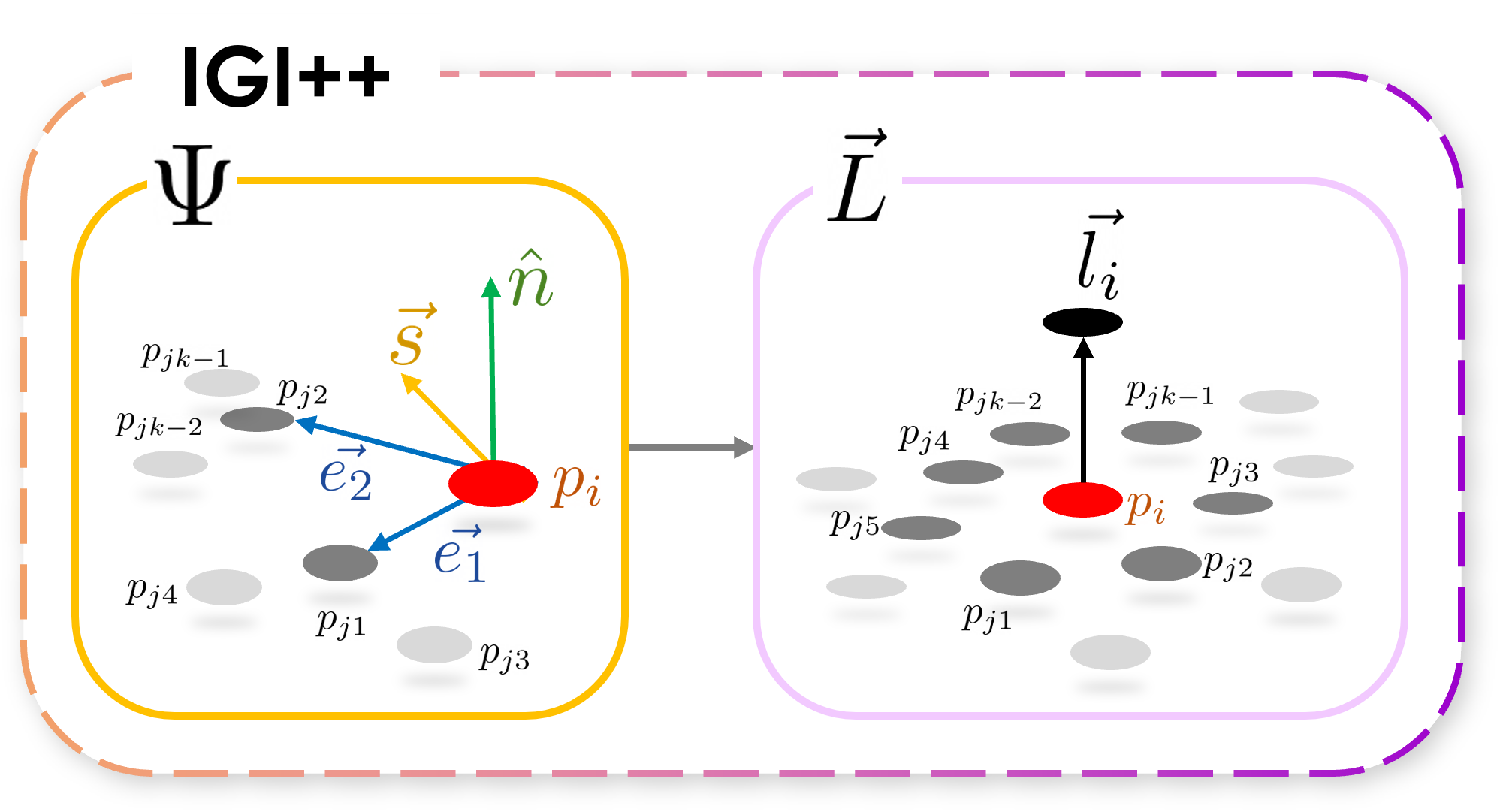}
            \caption {Illustration of proposed framework Intrinsic Geometry Interpreters++ (IGI++). \textbf{Left}: $\Psi$ depicts computation of IGI features (normal $\hat{n}$, std vector $\vec{s}$, and edge vectors $\vec{e_1}, \vec{e_2}$) using query point $p_i$ and two nearest neighbour $p_{j1}, p_{j2}$. \textbf{Right}: depicts computation of Laplace vectors $\vec{L}$ using query point $p_i$ and neighbour points $p_{j1\to k-1}$.}
            \label{fig:igipp}
        \end{figure}

        The FSL hypothesis of our network is notably aided by learning the basic local geometric interpreters IGI++ that incorporate Intrinsic Geometry Interpreters $\Psi$ and Laplace vectors $\vec{L}$. Following IGI which is proposed by VG-VAE~\cite{Vg-vae}, we introduce IGI++ and introduced group deviation vectors to facilitate the extraction of essential topological features as shown in Figure \ref{fig:igipp}. Laplace vectors play a crucial role in capturing abstract information related to edges and corners of point clouds, which are essential for few-shot learning.

        \noindent \textbf{Intrinsic Geometry Interpreter} $\Psi$ is a set of basic local geometric features that are fast to compute and capture the local intrinsic topology of the point cloud. $\Psi$ of a point cloud $P$ is given by:
        \begin{equation}
        \label{eq:igi}
            \displaystyle
            \Psi = \begin{cases}
                        p_{i} = {x,y,z};  & p_{i} \in \mathbb{R}^{3},\\
                        \vec{e_{1}} = \frac{p_{j1} - p_{i}}{||\vec{e_{1}}||_{2}}; & \vec{e_{1}} \in \mathbb{R}^{3},\\
                        \vec{e_{2}} = \frac{p_{j2} - p_{i}}{||\vec{e_{2}}||_{2}}; & \vec{e_{2}} \in \mathbb{R}^{3},\\
                        \hat{n} = \vec{e_{1}} \times \vec{e_{2}}; & \hat{n} \in \mathbb{R}^{3}.\\
                        \vec{s} = std(p_{j}); & \vec{s} \in \mathbb{R}^{3}.\\
                    \end{cases}
        \end{equation}
        
        where ($\overrightarrow{e_{1}}, \overrightarrow{e_{2}}$) represents edge (relative positions) (1, 2) respectively of a given point (position vector) $p_i$, ($|\overrightarrow{e_{1}}|, |\overrightarrow{e_{2}}|$) represents edge lengths, $\hat{n}$ represent normals of point cloud $p_{i}$ and $\vec{s}$ represent group deviation vector as illustrated in Figure \ref{fig:igipp}. Our proposed $\Psi \in \mathbb{R}^{15}$  captures a superior geometric features due to group deviation vector $\vec{s}$, unlike Enrich Features~\cite{Enrich-features} where $\Psi \in \mathbb{R}^{14}$. The relative positions and normals along with the position vector facilitate to capture of local geometric information that aids point cloud FSL. However, to capture abstract information such as edges and corners, we propose  Laplace vectors, which extract high-frequency information to further improve our network's performance. \\

        \noindent \textbf{Laplace Vectors} $\vec{L}$ are simple yet effective geometric signatures that capture the distribution, magnitude, and direction of the query to group deviation that facilitates extracting abstract information of edges and corners in a point clouds as shown in Figure\footnote{We acknowledge Ruben Wiersma for assisting us with these vector renderings.} \ref{fig:gpr}. Laplace vectors are given by:
         \begin{equation}
         \label{eq:LV}
             \vec{L} =  p_{i} \bigoplus \frac{1}{K}\sum_{j=0}^{K}(p_{j}-p_{i})
         \end{equation}

         where $p_{j}$ are local points of $p_{i}$ in a given k-NN (k-Nearest Neighbors) and  $p_{i} \in \mathbb{R}^{3}$, $\vec{L} \in \mathbb{R}^{6}$ and $\oplus$ is concatenation operation. The Illustration of Laplace vectors is depicted in Figure \ref{fig:igipp}. Laplace vectors allow us to operate with lower point density since they capture changes in local neighbourhoods. We propose to extract 30-dimensional Laplace vectors of 15-dimensional $\Psi$, i,e. $\vec{L}(\Psi)$ capturing superior geometric features as shown in Figure \ref{fig:teaser} towards facilitating Point Cloud FSL.

\subsubsection{Metric Learning}
    \label{sec:metrci_learning}

    Metric learning is essential for few-shot learning since it enables the computation of similarity metrics with limited labeled data~\cite{Prototypical_Network}\cite{relational-Networks}. To achieve this, we aim to learn a distance metric that can robustly compare the similarity between examples. In contrast to Euclidean spaces, hyperbolic spaces offer unique properties, such as exponential growth of volume with distance, that allow for more effective modeling of complex hierarchical structures. To achieve this, we propose using the Poincar\'e ball model to embed features in hyperbolic spaces and perform efficient computations such as distance measurements and gradient updates. This is particularly useful in few-shot learning scenarios with limited labeled data and complex hierarchical structures~\cite{HIE}. Hyperbolic distance considers hierarchical relations between support and query examples, leading to improved discrimination, as shown in Figure~\ref{fig:gpr}. The distance between two points $x,y$ with curvature $\kappa$ in Hyperbolic (Poincar\'e) manifolds
  $\mathbb{P}_{\kappa}$ is given by: ~\begin{equation}
        \label{eq:pdis}
        \resizebox{0.9\hsize}{!}{$
            d_{\mathbb{P}_{\kappa}} (x,y) = \frac{1}{\sqrt{-\kappa}} arcosh \Bigg(\frac{-2\kappa\left\lVert x-y\right\rVert_{2} }{(1+ \kappa\left\lVert x\right\rVert_{2}) (1+ \kappa\left\lVert y\right\rVert_{2})} +1\Bigg)$}
    \end{equation}

    \subsection{Prototypical Classification}
        \label{sec:pc}
        We transform support and query point clouds into geometric feature vectors by $\vec{L}(\Psi(p^{i}))$ given by Equation \ref{eq:igi} and~\ref{eq:LV}. The  geometric features are aggregated using $\mathcal{A}$ a symmetric operation (max, mean and, sum) to get a permutation invariant global feature. The invariant global feature is used as inputs to a single fully connected MLP $f_{\theta}$, which makes predictions on the label of the query point cloud by computing metric/embedding distance $d(,)$ between prototypes $\mu$ and query embedding $f_{\theta} \big( \vec{L}(\Psi(p^{i}_{q}) \big)$, where $\mu$ represents the mean of support embedding $f_{\theta} \big( \vec{L}(\Psi(p^{i}_{s}) \big)$ as depicted in Algorithm \ref{alg:fsl}. To ensure that the predictions are accurate, we normalize the results into a probability distribution \cite{3DFSL-Seg} and calculate the cross-entropy loss~\cite{crossentropy} between this distribution and the actual ground truth labels. This process facilitates the optimization of the network and improves its ability to perform few-shot learning on point clouds.

            \newcommand\mycommfont[1]{\footnotesize\ttfamily\textcolor{blue}{#1}}
            \SetCommentSty{mycommfont}
            
            \SetKwInput{KwInput}{Input}                
            \SetKwInput{KwOutput}{Output}
            
            \begin{algorithm}[ht]
            \DontPrintSemicolon
            \caption{Training episode loss computation for prototypical point cloud networks}
              \label{alg:fsl}
              
              \KwInput{$\mathcal{D} = \{(p_{1},y_{1}),..., (p_{N},y_{N})\}$, $d(.)$ }  
              \tcc{where each $y_{i} \in \{1,...,K\}$ and $d(.)$ is distance metric (Euclidean or Hyperbolic) as mentioned in Eq.(\ref{eq:pdis})}
              \KwOutput{The loss $J$ for a randomly generated training episode.}
        
              $V \gets$ RANDOMSAMPLE(\{1,...,K\}, $N_C$)\;
              \For{$k=1$ \KwTo $K$}{   
                $\mathcal{S}_{k} \gets$  RANDOMSAMPLE ($\mathcal{D}_{k}$, $N_\mathcal{S}$)\;
                $\mathcal{Q}_{k} \gets$  RANDOMSAMPLE ($\mathcal{D}_{k} / \mathcal{S}_{k}$, $N_\mathcal{Q}$)\;
                $\mu \gets \frac{1}{N_C} \sum_{(p_i,y_i) \in \mathcal{S}_k} f_\phi(p_{i})$\;  
                \tcc{ where $\mu$ is point cloud prototype and $f_\phi() = f_{\theta}(\vec{L}(\Psi))$ as mentioned in Eq.(\ref{eq:igi}) and Eq.(\ref{eq:LV})} 
                }
                
               $J \gets 0$
              \For{$k=1$ \KwTo $K$}{   
                    \For{ ($p,y$) in $\mathcal{Q}_k$}{
                    $J \gets J + \frac{1}{N_C N_\mathcal{Q}} \Big[ d(f_\phi(p),\mu) + \log{\sum_{k'}\exp{-d(f_\phi(p),\mu)}}\Big]$
                }}

            \end{algorithm}




\useunder{\uline}{\ul}{}
\begin{table*}[ht]
\caption{Few-shot classification results on ModelNet40~\cite{modelnet} dataset. With only 512 points, our method GPr-Net (Hyp) achieves state-of-the-art accuracy on  5-way 10-shots and 10-way 10-shots, where Hyp represents Hyperbolic and Euc represnts Euclidean distance metric in Algorithm \ref{alg:fsl}. Additionally, we have provided the parameters and performed Forward/Backward pass evaluations for the methods their source code was publicly available, using PyTorch-summary for a batch size of 150 indicating a 5-way 10-shot 20-query setting. The quantitative results (accuracies in $\%$) are represented in three styles: (1) \uline{\textbf{best}}, (2) \textbf{second best}, (3) \uline{third best} and $\pm$ represents mean and standard deviation of 6 experiments with a different random seed.
}
\label{tab:SOTA_MN40}
\resizebox{\textwidth}{!}{%
\begin{tabular}{rccccccc}
\Xhline{4\arrayrulewidth}
                           &      & \multicolumn{2}{c}{\textbf{5-way}}  & \multicolumn{2}{c}{\textbf{10-ways}} &                      &                     \\ \cline{3-6}
\multirow{-2}{*}{\textbf{Method}} &
  \multirow{-2}{*}{\textbf{Num Points}} &
  \cellcolor[HTML]{EFEFEF}\textbf{10-shots} &
  \cellcolor[HTML]{C0C0C0}\textbf{20-shots} &
  \cellcolor[HTML]{EFEFEF}\textbf{10-shots} &
  \cellcolor[HTML]{C0C0C0}\textbf{20-shots} &
  \multirow{-2}{*}{\textbf{\#Params}} &
  \multirow{-2}{*}{\textbf{F/B pass}} \\ \hline
\textbf{3D-GAN}~\cite{3D-GAN}            & 1024 & 55.80 $\pm$ 10.68 & 65.80 $\pm$ 09.90      & 40.25 $\pm$ 06.49       & 48.35 $\pm$ 05.59      & -                    & -                   \\
\textbf{Latent-GAN}~\cite{Latent-GAN}        & 1024 & 41.60 $\pm$ 16.91 & 46.20 $\pm$ 19.68      & 32.90 $\pm$ 09.16       & 25.45 $\pm$ 09.90      & -                    & -                   \\
\textbf{PointCapsNet}~\cite{PointCapsNet}      & 1024 & 42.30 $\pm$ 17.37      & 53.00 $\pm$ 18.72      & 38.00 $\pm$ 14.30       & 27.15 $\pm$ 14.86      & 2.15M                & 39GB                \\
\textbf{FoldingNet}~\cite{FoldingNet}        & 1024 & 33.40 $\pm$ 13.11      & 35.80 $\pm$ 18.19      & 18.55 $\pm$ 06.49       & 15.44 $\pm$ 06.82      & 0.67M                & 5.7GB               \\
\textbf{PointNet++}~\cite{pointnetpp}        & 1024 & 38.53 $\pm$ 15.98      & 42.39 $\pm$ 14.18      & 23.05 $\pm$ 06.97       & 18.80 $\pm$ 05.41      & 1.48M                & 149GB               \\
\textbf{PointCNN}~\cite{PointCNN}          & 1024 & 65.41 $\pm$ 08.92      & 68.64 $\pm$ 07.00      & 46.60 $\pm$ 04.84       & 49.95 $\pm$ 07.22      & -                & -                   \\
\textbf{PointNet}~\cite{pointnetpp}          & 1024 & 51.97 $\pm$ 12.17      & 57.81 $\pm$ 15.45      & 46.60 $\pm$ 13.54       & 35.20 $\pm$ 15.25      & 3.47M                & 8.5GB               \\
\textbf{DGCNN}~\cite{dgcnn}             & 1024 & 31.60 $\pm$ 08.97      & 40.80 $\pm$ 14.60      & 19.85 $\pm$ 06.45       & 16.85 $\pm$ 04.83      & 1.82M                & 53GB                \\
\textbf{SS-FSL (PointNet)}~\cite{SS-FSL} & 1024 & 63.20 $\pm$ 10.72      & 68.90 $\pm$ 09.41      & 49.15 $\pm$ 06.09       & 50.10 $\pm$ 05.00      & 3.47M                & 8.5GB               \\
\textbf{SS-FSL (DGCNN)}~\cite{SS-FSL}    & 1024 & 60.00 $\pm$ 08.87      & 65.70 $\pm$ 08.37      & 48.50 $\pm$ 05.63       & 53.00 $\pm$ 04.08      & 1.82M                & 53GB                \\
\textbf{Enrich-Features}~\cite{Enrich-features} &
  1024 &
  \cellcolor[HTML]{ECF4FF}{\ul 76.69 $\pm$ NA} &
  \cellcolor[HTML]{CDE1FF}{\ul \textbf{85.76 $\pm$ NA}} &
  \cellcolor[HTML]{ECF4FF}{\ul 68.76 $\pm$ NA} &
  \cellcolor[HTML]{CDE1FF}{\ul \textbf{80.72 $\pm$ NA}} &
  - &
  - \\ \Xhline{4\arrayrulewidth}
\textbf{GPr-Net (Euc)}        & 1024 & 74.37 $\pm$ 02.00 & 75.12 $\pm$ 02.08 & 62.14 $\pm$ 01.91  & 63.43 $\pm$ 02.05 & {\ul \textbf{1.24K}} & {\ul \textbf{50KB}} \\
\textbf{GPr-Net (Hyp)} &
  1024 &
  \cellcolor[HTML]{DAE8FC}\textbf{80.40 $\pm$ 00.55} &
  \cellcolor[HTML]{ECF4FF}{\ul 81.99 $\pm$ 00.91} &
  \cellcolor[HTML]{DAE8FC}\textbf{70.42 $\pm$ 01.80} &
  \cellcolor[HTML]{ECF4FF}{\ul 72.83 $\pm$ 01.78} &
  {\ul \textbf{1.24K}} &
  {\ul \textbf{50KB}} \\ 
\textbf{GPr-Net (Euc)}        & 512  & 74.04 $\pm$ 02.33 & 74.98 $\pm$ 02.42 & 62.31 $\pm$ 02.01  & 63.33 $\pm$ 02.21 & {\ul \textbf{1.24K}} & {\ul \textbf{50KB}} \\
\textbf{GPr-Net (Hyp)} &
  512 &
  \cellcolor[HTML]{CDE1FF}{\ul \textbf{81.13 $\pm$ 01.51}} &
  \cellcolor[HTML]{DAE8FC}\textbf{82.71 $\pm$ 01.28} &
  \cellcolor[HTML]{CDE1FF}{\ul \textbf{71.59 $\pm$ 01.16}} &
  \cellcolor[HTML]{DAE8FC}\textbf{73.78 $\pm$ 01.99} &
  {\ul \textbf{1.24K}} &
  {\ul \textbf{50KB}} \\ \Xhline{4\arrayrulewidth}

\end{tabular}%
}
\end{table*}

\section{Experiments}
\label{sec:Exp}
In this section, we investigate the effectiveness of the topological point embeddings generated by our classifier $f_{\theta}$ for few-shot 3D object classification, using the dataset ModelNet40 \cite{modelnet}. ModelNet40 dataset encompasses 40 object categories that include a collection of 12,311 models. Note that our model is trained on an Nvidia GTX 1050ti GPU and PyTorch 1.11 and we use geoopt \cite{geoopt} for the hyperbolic operations. 

\subsection{Few-Shot 3D object classification} 
    We evaluate the impact of the proposed IGI++ in our network for point cloud few-shot learning. We report the mean and standard deviation of our results with 95\% confidence scores across 6 experiments with different seeds towards better reproducibility in Table~\ref{tab:SOTA_MN40}. Unlike Enrich Features \cite{Enrich-features} and SS-FSL~\cite{SS-FSL} our model is trained and tested in a few-shot setting, using only the coordinates ($x,y,z$) of each point. To compute Laplace vectors the number of k for the nearest neighbors is set to k = 40. The 30-dimensional Laplace vector is mapped to a 32-dimensional point using single MLP $f_{\theta}$ as explained in Section~\ref{sec:pc}. We use the SGD (Stochastic Gradient Descent) optimizer for Euclidean metric and RSGD~\cite{geoopt} for hyperbolic metric, with a momentum of 0.9 and weight decay of 0.0001, with the learning rate reduced from 0.1 to 0.001 through the cosine annealing in Algorithm~\ref{alg:fsl} for 50 epochs of 4 train  and 300 test few-shot episodes. We ensure the FSL paradigm via maintaining $D_{train} \cap D_{test} = \emptyset$ such that, for each experiment, we randomly sample $\mathcal{T}$ categories of data to form $D_{train}$ and rest categories without replacement form $D_{test}$ this meta-training strategy aids in understanding true robustness of the proposed method in FSL. For our experiments, we randomly sampled $\mathcal{T}=24$ categories for training and 16 for testing in ModelNet40 as suggested by~\cite{Enrich-features}. Note that the results  of all other networks except Enrich-Features~\cite{Enrich-features} in Table~\ref{tab:SOTA_MN40} are derived from SS-FSL~\cite{SS-FSL}.

    \begin{figure*}
        \centering
        \includegraphics[width=\textwidth]{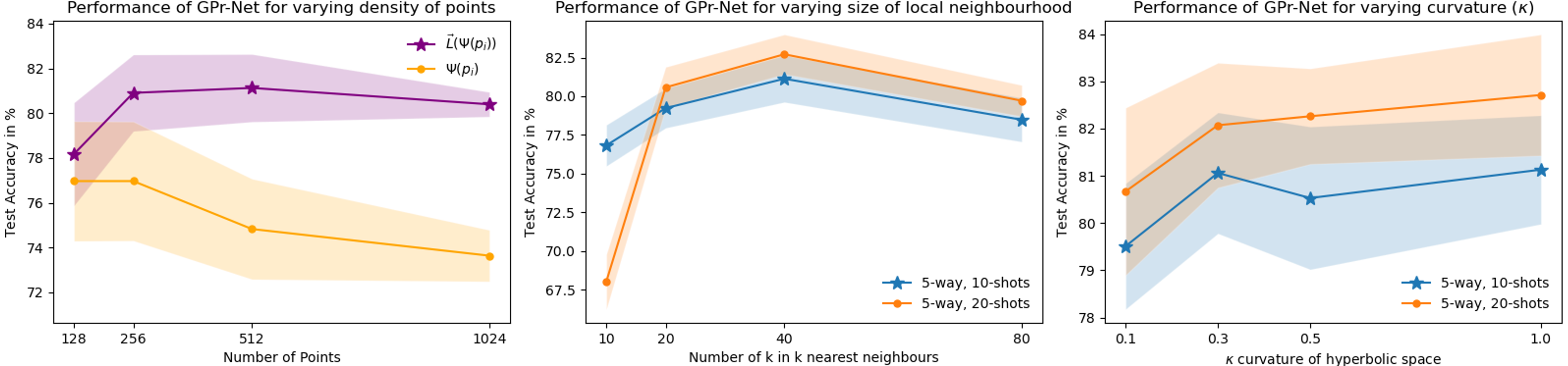}
        \caption{The \textbf{Left}: compares the few-shot accuracy of GPr-Net with and without Laplace vectors $\vec{L}$ as the point density varies. The results indicate that incorporating Laplace vectors leads to superior performance. The \textbf{Center}: impact of k-Nearest Neighbors (k-NN) is analyzed by varying k on GPr-Net with Laplace vectors. It is observed that k=40 yields the best results for 5-way 10 and 20-shots. Finally, the \textbf{Right}: investigates the effect of hyperbolic curvature on few-shot accuracy. GPr-Net with $\kappa \rightarrow 1$ hyperbolic metric outperforms the Euclidean metric with $\kappa \rightarrow 0$ for both 5-way 10 and 20-shots. Further details are provided in Section \ref{sec:ablation}.}
        \label{fig:ablation}
    \end{figure*}

    \subsection{Comparison with State-of-the-art Methods}

        Our novel GPr-Net not only outperforms existing methods that rely on data-heavy pre-training tasks like SS-FSL~\cite{SS-FSL} or complex feature extractors~\cite{Enrich-features}\cite{cia} in terms of accuracy, but it also operates much faster. To demonstrate this, we compared our proposed backbone architecture to several open-source backbones and evaluated parameters, few-shot classification accuracy, and parameters on the ModelNet40 dataset~\cite{modelnet}, as suggested by PointMLP~\cite{PointMLP}. For example, SS-FSL (DGCNN) is a cumbersome model that achieves impressive results with 1.82M parameters and a forward/backward pass of 53GB, as shown in Table~\ref{tab:SOTA_MN40}. In contrast, our GPr-Net achieves state-of-the-art FSL accuracy on point clouds while maintaining only 1.24K parameters, which is 280 times less than SS-FSL, and a forward/backward pass of 50KB. This is particularly essential for applications like robotics, self-driving cars, and others that require deploying these models efficiently.
        
        Our results with Hyperbolic-metric, presented in Table~\ref{tab:SOTA_MN40}, demonstrate that our method surpasses Enrich-Features~\cite{Enrich-features} by 5\% for 5-way 10-shots and 3\% for 10-way 10-shots. However, when compared to SS-FSL~\cite{SS-FSL}, our method achieves significant improvements of 18\% for 5-way 10-shots, 14\% for 5-way 20-shots, 22\% for 10-way 10-shots, and 21\% for 10-way 20-shots. These results highlight the efficacy of our approach in addressing the challenging problem of few-shot learning for point clouds. \textit{Notice that we achieve this by only 50 epochs, 4 training episodes and only 512 points}. Remarkably, our method achieves the smallest standard deviation across 6 experiments with different random seeds, indicating its robustness and lack of bias towards a particular category.

    \subsection{Ablation Studies}
        \label{sec:ablation}
        This section presents ablation studies to analyse the impact of different designs of our proposed module on few-shot point cloud classification.\\
        
        \noindent \textbf{Significance Laplace Vectors} $\vec{L}$ is demonstrated in the left of Figure~\ref{fig:ablation}. The results show that the GPr-Net with $\vec{L}$ outperforms the one without in all cases of point density variation, as reported by the mean and standard deviation accuracy of 6 experiments for 5-way 10-shot tasks on the Hyperbolic variant of GPr-Net. The experiments were conducted with 128, 256, 512, and 1024 points, and the superior performance of the classifier with $\vec{L}$ suggests the significance of extracting geometric features using Laplace vectors for effective few-shot learning on point clouds.
        
        \noindent \textbf{Local Neighbourhood Size} justifies the effectiveness of the Laplace vectors in our proposed GPr-Net for few-shot learning on point clouds, we conducted additional experiments to determine the appropriate value of k in k-NN for computing Laplace vectors. We aim to determine the best value of k that describes local topological changes such as edges and corners, which are largely dependent on the size of the local neighborhood. In the center of Figure~\ref{fig:ablation}, we present our findings on the need for selecting an appropriate value of k. We report the mean and standard deviation accuracy of 6 experiments for 5-way 10-shot and 5-way 20-shot tasks on the Hyperbolic variant of GPr-Net. The experiments were conducted with k=10,20,40 and 80 for 512 points in a point cloud. Our findings suggest that k=40 is the optimal value for 512-point density in both 10 and 20-shot cases.

        \noindent \textbf{Influence of Curvature $\kappa$} plays a critical role in the performance of hyperbolic metrics for point cloud few-shot learning. The negative curvature allows for more efficient space utilization, increasing the ability to distinguish between points. We perform experiments that show an increase in curvature $\kappa$ in Eq.~\ref{eq:pdis} results in improved performance for few-shot learning tasks, as the embeddings are better able to capture the similarities and differences between point clouds. However, as $\kappa$ approaches zero, the hyperbolic space approaches a Euclidean space, and the benefits of the negative curvature are lost as Depicted in the right of Figure~\ref{fig:ablation}. Therefore, it is essential to find the optimal curvature for a given few-shot learning task to achieve the best results. Our findings indicate that $\kappa = 1.0$ is the best suited for 5-way 10-shots and 5-way 20-shots tasks for 512 points with Laplace vectors and k = 40.

        \useunder{\uline}{\ul}{}
        \begin{table}[ht]
        \centering
        \caption{The comparison of our proposed GPr-Net with state-of-the-art methods CGNN~\cite{cgnn} and CIA~\cite{cia} in a 5-way 1-shot and 5-way 5-shot learning setting. The results demonstrate the competitive performance of our method with significantly fewer parameters and faster training speed. Although CIA~\cite{cia} achieves state-of-the-art accuracy in both settings, our method achieved the second-best result, trained with only 4 episodes.}
        \label{tab:5w1s}
        \begin{tabular}{rcc}
        \Xhline{2\arrayrulewidth}
        \multirow{2}{*}{\textbf{Method}} & \multicolumn{2}{c}{\textbf{5-way}}                        \\ \cline{2-3} 
                                         & \textbf{1-shot}             & \textbf{5-shots}             \\ \Xhline{2\arrayrulewidth}
        \textbf{CGNN~\cite{cgnn}}                   & -                           & 76.85 $\pm$ NA                       \\
        \textbf{CIA~\cite{cia}}                     & { \textbf{75.70 $\pm$ 0.74}} & {\textbf{87.15 $\pm$ 0.47}} \\ \hline
        \textbf{GPr-Net (Euc)}              & 64.12 $\pm$ 0.73             & 74.56 $\pm$ 1.03             \\
        \textbf{GPr-Net (Hyp)}              & \uline{67.91 $\pm$ 1.07}    & \uline{79.09 $\pm$ 0.97}    \\ \Xhline{2\arrayrulewidth}
        \end{tabular}%
        \end{table}
        
        \noindent \textbf{Performance Efficiency} of proposed method GPr-Net is conducted and we compare with CGNN~\cite{cgnn} and CIA~\cite{cia} in a 5-way 1-shot and 5-way 5-shot learning setting. Our method was trained for only 4 episodes, and we observed competitive performance with the other two models, as shown in Table~\ref{tab:5w1s}. CGNN~\cite{cgnn} and CIA~\cite{cia} also aim to learn representations and relations between prototypes and query features by utilizing feature-matching methods such as graph neural networks or self-attention mechanisms, respectively. Although CIA~\cite{cia} achieves state-of-the-art accuracy in 5-way, 1-shot and 5-shot settings, we still achieved the second-best result with significantly fewer parameters and faster training speed. Unfortunately, since the code for CIA~\cite{cia} was not open-source, we were unable to make a direct comparison in terms of speed.

    \subsection{Embedding Visualization}

\begin{figure}[ht]
            \center
            \includegraphics[width=1.0\linewidth]{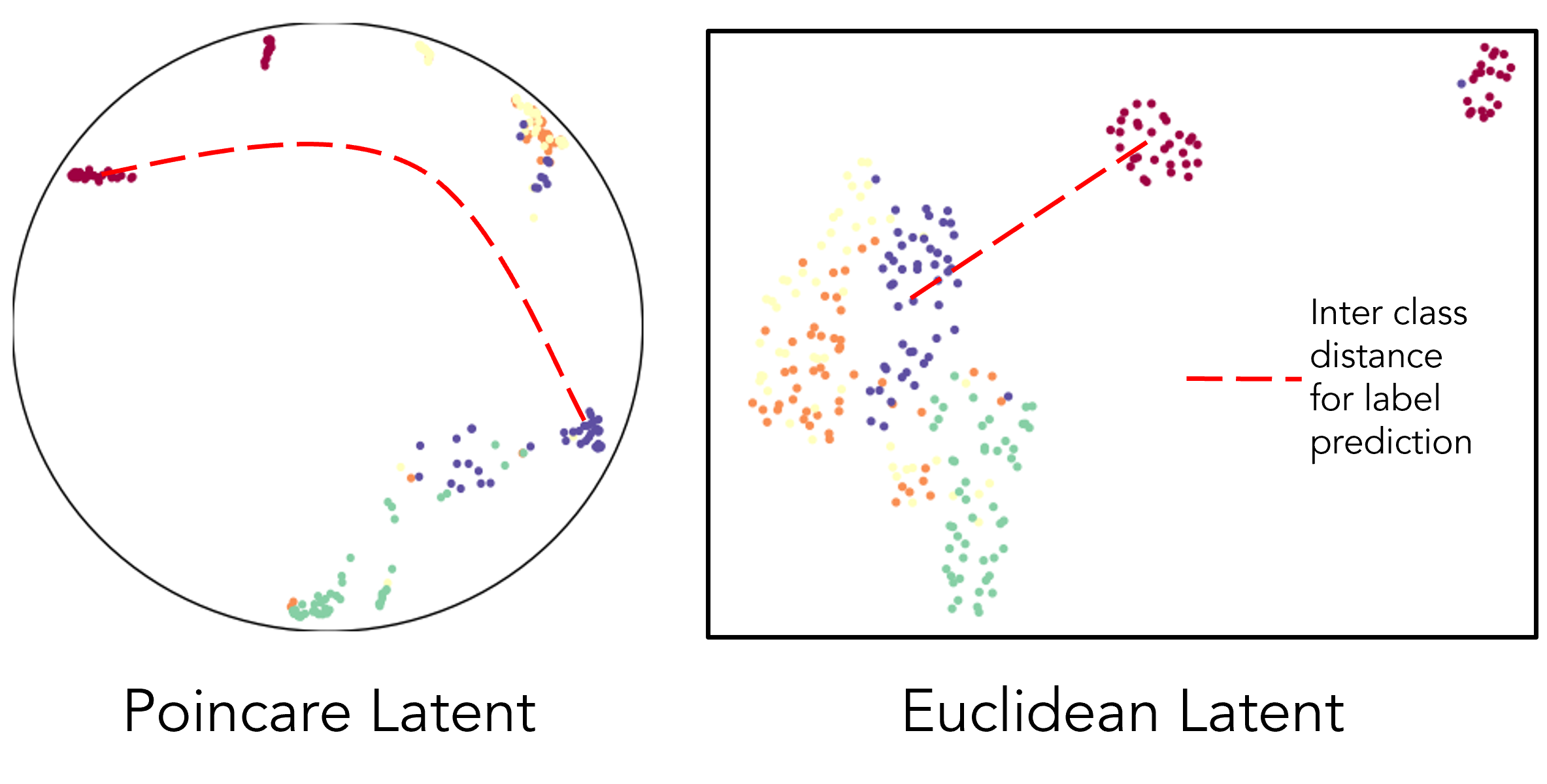}
            \caption{A visualization of the embeddings learned for the few–shot task. \textbf{Left}: Our 5-way task on Modelnet40 with Poincar\'e metric. \textbf{Right}: Our 5-way task on Modelnet40 with Euclidean metric. The two-dimensional projection was computed using the UMAP~\cite{umap-software}.}  
            \label{fig:latent} 
        \end{figure}

        In the context of a \textit{5way-10shot-50query} setting on the ModelNet40~\cite{modelnet} dataset, we present a visualization of the features generated by our proposed GPr-Net. Specifically, we compare the features obtained using the hyperbolic/Poincar\'e metric with those obtained using the Euclidean metric. The left-hand side of Figure~\ref{fig:latent} corresponds to the former, while the right-hand side depicts the latter.
        
        It is worth noting that there exists a significant distribution shift between the support and query features in this setting. However, our proposed hyperbolic metric helps to mitigate the challenge of inter-class ambiguity by causing the features to move towards the boundary of the Poincar\'e manifold. This allows for better differentiation between classes, leading to more accurate predicted labels.

    \subsection{Limitations}
        Notwithstanding the promising results achieved in our study, we acknowledge certain limitations that need to be addressed. First, due to the nature of our approach incorporating Hyperbolic Projection, it is currently hard to perform part segmentation on per-point embeddings. This is a known limitation also faced by other works in the field, such as HyCoRe~\cite{HyCoRe}. As a result, we were unable to apply our method to datasets that require part segmentation, limiting the scope of our study. Another limitation of our method is susceptibility towards noise due to the use of k-NN for local grouping. Therefore, we exclude certain real-world datasets, such as the Sydney and ScanObjectNN datasets, which contain large amounts of noise.
        
        Despite these limitations, we believe that our study offers valuable insights into few-shot learning in point cloud settings. We hope that our findings will inspire further research to address these limitations and lead to the development of more robust and effective methods for few-shot learning in point clouds.
        
\section{Conclusions}
\label{sec:Conc}
    In this work, we have proposed a new perspective on point cloud few-shot learning by challenging the assumption of complex network designs and training strategies. Our proposed lightweight Geometric Prototypical Network, GPr-Net, leverages simple yet effective geometric signatures, including the Intrinsic Geometry Interpreter and Laplace vectors, to efficiently capture the intrinsic topology of point clouds and achieve superior performance on the ModelNet40 dataset. Additionally, employing a hyperbolic/poincar\'e metric for learning 3D few-shot features further improves the effectiveness of our approach. Our experimental results demonstrate that GPr-Net outperforms state-of-the-art point cloud few-shot learning techniques with 5\% higher accuracy and $170\times$ fewer parameters than all existing works.

\section{Broader Impact}
    The focus of this study is on local geometric features and selecting appropriate metric space that can enhance the performance of few-shot learning in point cloud settings, even when labeled data is limited. 
    
    Point cloud few-shot learning has the potential to impact a wide range of fields. For example, robotics can accelerate the training of robots to recognize new objects with a small number of labeled examples, resulting in faster and more cost-effective learning. In architecture and engineering, it can facilitate the design of complex structures with limited labeled data by allowing shape analysis and thermal analysis. Additionally, it can assist in environmental studies, such as land surveying, by making it easier to classify and analyze 3D point cloud data with minimal labels.

{\small
\bibliographystyle{ieee_fullname}
\bibliography{main}
}

\end{document}